\pdfoutput=1

\documentclass[11pt]{article}

\usepackage{latex/acl}

\usepackage{times}
\usepackage{latexsym}

\usepackage[T1]{fontenc}

\usepackage[utf8]{inputenc}

\usepackage{microtype}

\usepackage{inconsolata}
\usepackage{tabularx}

\usepackage{booktabs}
\usepackage{graphicx}
\usepackage{ amssymb }

\usepackage{soul}

\usepackage{fixltx2e}

\usepackage{todonotes}

\usepackage{hyperref}

\usepackage{amsmath}
\usepackage[frozencache,cachedir=.]{minted}
\newcounter{todocounter}

\title{Explicating the Implicit: Argument Detection Beyond Sentence Boundaries}

\newcommand{\authorspace}{\hspace{9pt}}

\author{Paul Roit$^1$ \authorspace Aviv Slobodkin$^1$ \authorspace Eran Hirsch$^1$ \authorspace Arie Cattan$^1$\\
{\bf Ayal Klein$^1$ \authorspace Valentina Pyatkin$^{2,3}$ \authorspace Ido Dagan$^1$} \\
{$^1$Bar-Ilan University}\authorspace{$^2$Allen Institute for Artificial Intelligence}\\{$^3$University of Washington}\\
 \footnotesize{\texttt{plroit@gmail.com}}
} 

\begin{document}
\maketitle

\begin{abstract}
Detecting semantic arguments of a predicate word has been conventionally modeled as a sentence-level task. 
The typical reader, however, perfectly interprets predicate-argument relations in a much wider context than just the sentence where the predicate was evoked.
In this work, we reformulate the problem of argument detection through textual entailment to capture semantic relations across sentence boundaries.
We propose a method that tests whether some semantic relation can be inferred from a full passage by first encoding it into a simple and standalone proposition and then testing for entailment against the passage.
Our method does not require direct supervision, which is generally absent due to dataset scarcity, but instead builds on existing NLI and sentence-level SRL resources. 
Such a method can potentially explicate pragmatically understood relations into a set of explicit sentences. 
We demonstrate it on a recent document-level benchmark, outperforming some supervised methods and contemporary language models.
\end{abstract}

\section{Introduction}
\label{sec:intro}
\begin{figure}[t!]
    \centering
    \includegraphics[width=\columnwidth]{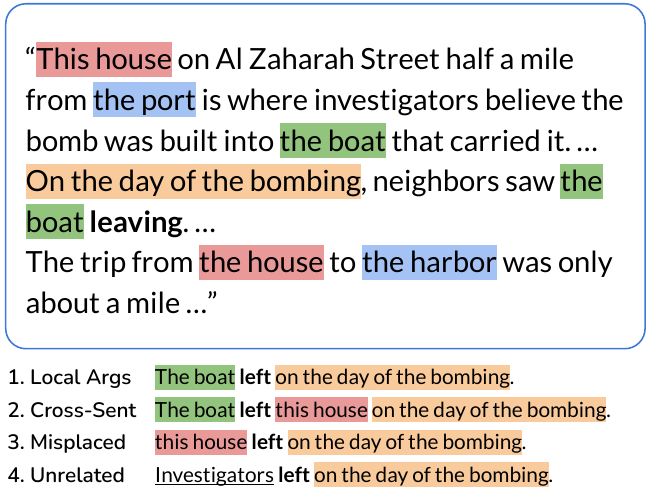}
    \caption{Example of semantic arguments in the sentence and document scope. The predicate is in boldface while arguments are highlighted in color. The bottom part shows four different propositions: (1) A proposition constructed from in-sentence arguments of the predicate. (2) The same proposition with additional arguments from other sentences in the document. (3) A proposition with some arguments (the house) placed in an incorrect syntactic position that does not align with its original semantic role. (4) A proposition with an incorrect argument according to the document. Both (3) and (4) are not supported by the document.}
    \label{fig:simple_example}
\end{figure}

Identifying which entities in a text play specific semantic roles with respect to a predicate word is a core ability of language comprehension \cite{Fillmore1976FRAMESA}.
Such basic semantic information is often surfaced via simple lexical and syntactical patterns in the sentence.
Readers however can perfectly interpret such semantic relations pragmatically in a wider context than a single sentence.  
Consider the example in \autoref{fig:simple_example}. 
The location that the boat left, `\textit{the house}', and the destination where it was headed to, `\textit{the port}', can be deduced by associating the trip with the leave event in the preceding sentence.
Such examples showcase where our semantic interpretation departs from syntax, and allow us to systematically investigate how humans and machines reason over events in text in cases where they cannot rely on easy-to-follow grammatical patterns.

In this work,\footnote{Our codebase, dataset, and models can be found at \url{https://github.com/plroit/semquest}} we address detecting cross-sentence semantic arguments for verbal and deverbal noun predicates.
We propose a method based on textual-entailment \cite{dagan-etal-2005-rte} and supervised only with NLI and sentence-level Semantics Role Labeling (SRL) \cite{gildea-jurafsky-2000-automatic} data.
It takes a document and a marked predicate and outputs a set of simple, easy-to-grasp sentences that incorporate semantic arguments from anywhere in the document (e.g. `the house' argument from a different sentence incorporated into the leave event in \autoref{fig:simple_example}, proposition 2).

We assume that an argument is omitted by the speaker from the predicate's sentence due to its redundancy in discourse while re-inserting it back into its designated position next to the predicate should not alter the meaning of the event in the passage.
Our basic idea is that a simple proposition constructed from a set of true arguments should be entailed from the passage (see props 1-2 in \autoref{fig:simple_example}), while any proposition that targets the same predicate and contains a non-argument phrase or a misplaced phrase should not be entailed (see props 3-4 in \autoref{fig:simple_example}).
Therefore, we design a method that starts at the local parse of the predicate, builds a proposition from the extracted in-sentence arguments, and then examines candidate phrases one by one from across the document by inserting them into different positions and testing for entailment.

Our method does not require a frame repository such as PropBank \cite{palmer-etal-2005-proposition} or FrameNet \cite{baker-etal-1998-berkeley-framenet} to operate.
Instead, it uses the explicit syntactic argument structure in the proposition as a syntactic surrogate \cite{michael-2023-case} for the underlying semantics of the predicate in the passage (see how the meaning changes in the misplaced argument example, prop 3 in \autoref{fig:simple_example}).

Some recent works from the event extraction literature apply similar slot-filling \cite{li-etal-2021-document} or entailment-based methods \cite{sainz-etal-2022-textual,lyu-etal-2021-zero}. 
However, they rely on a limited event ontology for predefined templates for argument extraction. 
In contrast, our work uses English syntax for creating propositions, akin to the clause structure in \citet{Corro2013ClauseIE}.

This illuminates another benefit of our approach, being schema-free, the propositions can be easily processed downstream by parsers trained on abundant single-sentence data, for example for relation extraction \cite{hendrickx-etal-2010-semeval} or event participant detection \cite{doddington-etal-2004-automatic}.
Thus, explicating to downstream tasks the set of document-level semantic relations that were previously unreachable, now encoded in a simple sentence form. 

The cross-sentence task has been notably under-explored in the literature, largely due to the extreme difficulties in constructing suitable datasets \cite{gerber-chai-2010-beyond,moor-etal-2013-predicate,ruppenhofer-etal-2010-semeval}.
Under this context, we suggest a significantly more feasible approach that leverages only existing resources designed for in-sentence argument detection to detect semantic arguments across sentence boundaries. 
Our distantly supervised method achieves higher performance than some fully supervised models on a document-level dataset \cite{elazar-etal-2022-text} for noun-phrase relations, and outperforms other zero and few-shot approaches on a re-annotated benchmark for verbal predicates \cite{moor-etal-2013-predicate}. 

\section{Background and Related Works}
\label{sec:background}
\textbf{Implicit Arguments} Mainstream research efforts in semantic role labeling (SRL) \cite{gildea-jurafsky-2000-automatic, kingsbury-palmer-2002-treebank} have focused on the problem of assigning semantic roles only to syntactically related phrases, e.g.\ the subject or object phrases of verbs, while neglecting constituents from the wider passage that are pragmatically interpreted as participants.
The latter ones, referred to as \textit{implicit} arguments \cite{gerber-chai-2010-beyond, ruppenhofer-etal-2010-semeval} despite being overtly understood by readers, constitute a sizeable portion of the potentially identified argument set \cite{fillmore1986pragmatically,gerber-chai-2010-beyond,klein-etal-2020-qanom,roit-etal-2020-controlled,pyatkin-etal-2021-asking}.
While some recent works \cite{fitzgerald-etal-2018-large} have annotated large datasets with semantic arguments captured anywhere within the sentence scope, to this day, only a handful of limited resources for SRL in the document scope exist \cite{gerber-chai-2010-beyond,moor-etal-2013-predicate,ruppenhofer-etal-2010-semeval,feizabadi-pado-2015-combining}.
Some resources contain only a few hundred instances, others lack diversity, capturing only a tiny set of predicates (5-10 unique verbs), and all focused only on semantic core roles (i.e. the numbered arguments in PropBank), neglecting other meaningful information for the reader such as temporal or locative modifiers.
\citet{ogorman-etal-2018-amr} annotated a dataset of cross-sentence arguments on top of AMR graphs \cite{banarescu-etal-2013-abstract} specifying arguments as AMR concepts, without their exact location in the sentence. 

Earlier supervised models for implicit SRL relied on extensive feature engineering and also using gold features \cite{gerber2012semantic}.
Many works additionally attempted to overcome data scarcity by creating artificial training data using coreference \cite{silberer2012casting} or aligning predicates in comparable documents \cite{roth2015inducing}. \citet{cheng-erk-2018-implicit} proposed to transform the problem into a narrative cloze task, creating synthetic datasets. 
More recently, \citet{zhang2020two} improved upon the baseline model proposed for the RAMS dataset \cite{ebner2020multi}, and trained a supervised model that detects argument heads before expanding to the full constituent.

\begin{figure*}[htp]
    \centering
    \includegraphics[width=\textwidth]{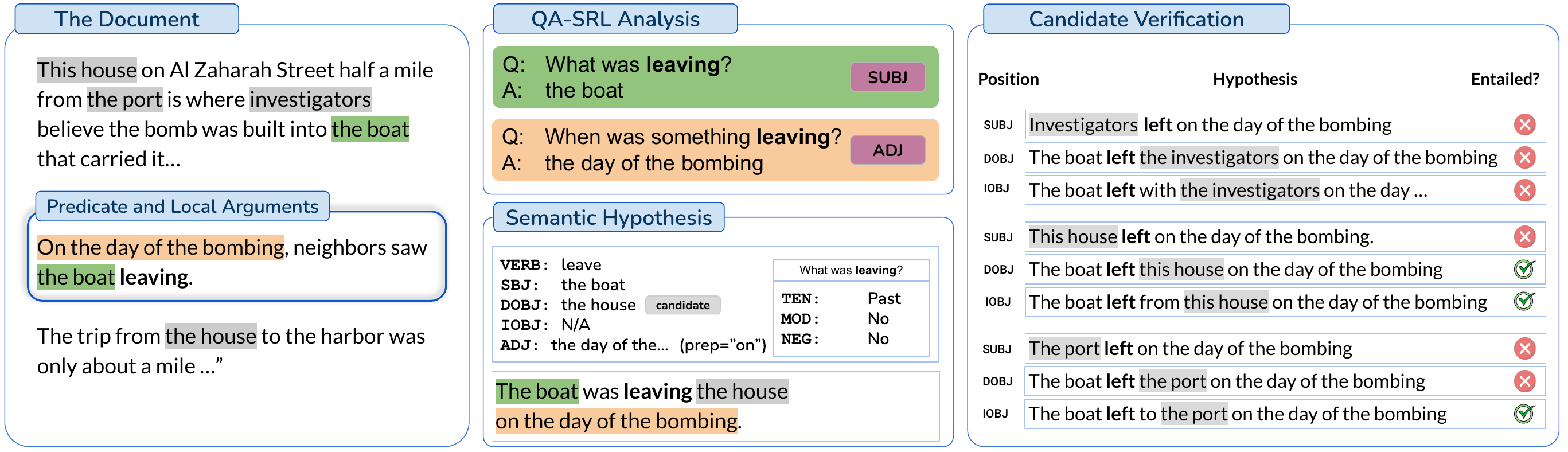}
    \caption{\textit{Left:} An excerpt from the document with the predicate marked in bold in the circled sentence. Candidate phrases from across the document are highlighted in gray. For completeness, we also highlight local arguments and their co-referent mentions in color. 
    \textit{Center-Top:} A QA-SRL parser analyzes the predicate's sentence and outputs local arguments, their questions, and their syntactic position (in purple).
    \textit{Center-Bottom:} Hypothesis fields are assigned with argument and candidate phrases to different syntactic positions. Grammatical attributes are extracted from the question of the first local argument.
    The generated hypothesis sentence is shown in the box below.
    \textit{Right:} Each candidate (highlighted in gray) is inserted into three different position fields and the resulting hypothesis is verified with an NLI model against the full document. The second candidate demonstrates two correct alternations. }
    \label{fig:method}
\end{figure*}

\textbf{QA-SRL} \cite{he-etal-2015-question} represents the label of each semantic argument as a simple Wh-question that the argument answers, for example, \textit{`Who acquired something?'} encodes the agent, and \textit{`Who did someone give something to?'} encodes the recipient.
These question-labels point at the syntactic position of the argument in a declarative form of the QA pair, e.g.: `The agent acquired something' or `Someone gave something to the recipient' (see the example in \autoref{fig:method}, top-left, where the position of the answer is apparent from the question).
Each question also encodes the tense of the event, the modality, and negation properties (might the event occur or has the event occurred?) which are used to instantiate our propositions. 
\citet{klein-etal-2020-qanom} extended QA-SRL to deverbal nominal predicates, recently leveraged for training a joint verbal and nominal QA-SRL model \cite{klein-etal-2022-qasem}. And \citet{pyatkin-etal-2020-qadiscourse} used QA pairs to represent implicit and explicit discourse relations, also across sentences \cite{pyatkin-etal-2023-design}.     

\textbf{TNE} is a dataset for modeling semantic relations between noun phrases (NPs) across a document and is annotated on top of Wikinews.
A relation consists of an anchor and complement phrases that are labeled with a preposition, i.e.\ [the investigation]\textsubscript{\textsc{Anchor}} \textit{by} [the police]\textsubscript{\textsc{Complement}}.
Each document is first segmented into a list of non-overlapping NPs and every NP pair is annotated with either a preposition or a 'no-relation' tag.
Each NP is also assigned to a cluster of co-referring within-document mentions.

\textbf{ON5V} \cite{moor-etal-2013-predicate} is a dataset derived from news articles in the development and train partitions of OntoNotes \cite{pradhan-xue-2009-ontonotes}.
The dataset contains 390 instances selected from 260 documents, and annotates \textit{five} different verbal predicates.
The original annotators have inspected only core roles (i.e. numbered: ARG0, ARG1, etc.) that were missing an explicit filler argument in OntoNotes, and filled the role with the closest phrase to the predicate that fit the role description.
We have re-annotated this dataset to close the coverage gap for modifier roles and retrieve all fitting phrases (see \S\ref{sec:datasets}). 

\section{Task Definition}

Given a text document $\mathbf{D} = [\mathbf{s}_1, \ldots, \mathbf{s}_{n}]$, consisting of $n$ sentences, and a target predicate word ${\mathrm{p}}$ from the document (a verb or deverbal noun), our task is to detect the set of semantic arguments ${\mathcal{A}_{\mathrm{p}} = \{a_i\}}$ of the predicate.
In this work, we adopt the QA-SRL definition of a semantic argument and apply it to the document-level task. An argument of $\mathrm{p}$ is any phrase $a$ in the document that pertains to the event referred by $\mathrm{p}$ and correctly answers a simple Wh-question revolving around the predicate word.
For example, \textit{What did something leave?} asks about \textit{the house} argument in \autoref{fig:simple_example}.

We will mark the sentence where the predicate resides as $\mathbf{s}_p$ and define the set of local, in-sentence arguments as ${\mathcal{L}_{\mathrm{p}} = \{ a_i | a_i \in \mathcal{A}_{\mathrm{p}} \land a_i \subset \mathbf{s}_p \}}$, where the subset notation is overloaded here to represent a sub-sequence.
While past works have focused on detecting $\mathcal{L}_\mathrm{p}$, we will show a method that leverages $\mathcal{L}_\mathrm{p}$ to detect the rest of cross-sentence arguments.

\section{Method}
\label{sec:method}

Our approach is based on synthesizing a simple sentence that encodes candidate phrases as arguments for the predicate $\mathrm{p}$, and testing if the sentence is entailed from the document.
The synthesized sentence ${\mathcal{H}}$, named the \textit{semantic hypothesis}, is created from templates by placing different phrases in subject or object positions around the predicate verb.\footnote{If $\mathrm{p}$ is a nominalization, its verbal form will be used instead.}
For example, in proposition no. 2 in \autoref{fig:simple_example} \textit{`The boat'} is assigned to the subject position and represents the \textsc{leaver} role, and \textit{`on the day of the bombing'} is assigned to an adjunct and represents a \textsc{temporal} role.
If the semantic hypothesis is entailed from the passage, we conclude that the predicate-argument relations encoded within $\mathcal{H}$ are manifested in the document.
On the other hand, if the hypothesis is not entailed, we can conclude that the document does not express one or more of the encoded argument relations.
For example, proposition no. 4 incorrectly assigns \textit{`the investigators'} to the \textsc{leaver} role, which should not be entailed from the passage.

Our method iterates over candidate phrases extracted from across the document (\autoref{fig:method}, left).
To determine if a candidate phrase $\mathrm{c}$ is a semantic argument of the predicate, we construct several hypotheses that incorporate the candidate and check for entailment against the document (\autoref{fig:method}, right).
The hypotheses combine local, in-sentence, arguments of the predicate with the candidate phrase, and differ in the syntactic position that the candidate assumes.
Conceptually, each position encodes the candidate into a different semantic role. 
Moreover, the hypothesis sentence associates the candidate with the specific target event by incorporating local arguments of the event, the remote candidate, and the target predicate into a single sentence.
If $\mathcal{H}$ is entailed, it suggests that the candidate participates in the event in the document, and that the candidate's semantic role in the document is the same as its role in $\mathcal{H}$.

Our synthetic hypothesis sentence is generated from syntactic templates.
They accommodate the main verb \texttt{(VERB)} with up to 4 argument phrases in the following unique syntactic positions: subject \texttt{(SUBJ)}, direct object \texttt{(DOBJ)}, indirect object \texttt{(IOBJ)}, and adjunct \texttt{(ADJ)}.
It can be modified using the following grammatical attributes: tense \texttt{(TEN)}, negation \texttt{(NEG)}, and modality \texttt{(MOD)}.
These fields are inspired by the QA-SRL question syntax that was similarly used to generate simple propositions in \citet{pyatkin-etal-2021-asking}.
They present high versatility and can capture any semantic role of a verbal predicate in English, yet maintain a compact representation requiring only a few assignments.

In this work, unless specified otherwise, we consider as candidates noun phrases and named entities from the document that do not overlap with local arguments, including those from $\mathbf{s}_p$. 
Each candidate is placed into different hypotheses in the subject, direct and indirect object positions.
We use the probability of entailment as predicted by an NLI model as the score of the hypothesis ${\mathrm{NLI}(\mathbf{D}, \mathcal{H})}$ and define the score of each candidate to be 
\[
 \max\limits_{\mathrm{pos} \in \{\texttt{S,D,I}\}} \mathrm{NLI}\left(\mathbf{D}, \mathcal{H}[\mathrm{pos}=c]\right)
\]
the maximal score of any of its hypotheses.
Our final predicted set of arguments includes all candidate phrases that scored above a configurable threshold and all local arguments predicted by the parser.
In the next subsections, we will describe how we populate the hypothesis syntactic positions and attributes given the predicate and its sentence (\S\ref{sec:structure_from_qasrl}), and how we then generate the actual hypothesis sentence (\S\ref{sec:semantic_hypothesis}).

\subsection{Populating Hypothesis Fields}
\label{sec:structure_from_qasrl}
We initialize the hypothesis fields with the predicate as the main verb and assign its local arguments to syntactic positions.
The local arguments are extracted from the predicate's sentence along with their question labels using a high-performing \mbox{QA-SRL} parser \cite{klein-etal-2022-qasem} (see \autoref{fig:method}, center-top).
The syntactic position of each argument is determined by applying a heuristic from \citet{klein-etal-2020-qanom} that inspects the question and maps it to one of the argument positions (\autoref{fig:method}, top-left).
Generally, each local argument is assigned to a field according to its syntactic position (\autoref{fig:method}, center-bottom).
However, to mitigate QA-SRL parsing errors and resolve collisions, we score each local argument, and select the top argument per position.
To score a local argument, we generate a singleton hypothesis containing only the argument, and compute its entailment probability against the predicate's sentence using an NLI model.
Refer to \S\ref{sec:impl_details} for more details.

Tense, modality, and negation attributes are determined by inspecting the question of the first local argument in the hypothesis.
For example, \textit{`Who might have left?'} indicates that the event should be described in the past tense with the modal verb `might' (see \S\ref{sec:impl_details}).

Finally, given a candidate phrase, we generate different hypotheses, assigning the candidate to a different syntactic position in each, possibly overriding a local argument.

\subsection{Generating the Semantic Hypothesis}
\label{sec:semantic_hypothesis}
Given an assignment to the hypothesis fields, we construct the hypothesis sentence by filling a syntactic template and using English grammar rules for inflection.
If a subject phrase is specified, we use the active voice template.
Otherwise, we resort to the passive voice, placing the direct object in the subject position as follows:
\begin{align*}
&\text{Active } \implies \texttt{SUBJ-AUX-VERB-DOBJ-IOBJ-ADJ} \\
&\text{Passive} \implies \texttt{DOBJ-AUX-VERB\textsubscript{passive}-IOBJ-ADJ}
\end{align*}

The corresponding auxiliary (\texttt{AUX}) and main verbs are automatically assigned or inflected\footnote{We use \url{https://github.com/bjascob/pyinflect}} based on the grammatical attributes of the hypothesis, the active or passive voice, and after considering the subject-verb agreement.

Notably, a valid declarative sentence in English must contain a subject phrase, while for some instances there is no apparent local subject or object argument.
E.g. `\textit{the 2 PM \textbf{presentation}}' implies that someone is presenting something at 2 PM. 
To write a valid hypothesis, we insert abstract placeholder arguments instead of concrete ones, placing `someone' in an empty subject position or `something' in empty object positions when necessary (see \S\ref{sec:impl_details} for details).

The prepositions for indirect objects or adjuncts are assigned by either syntactic analysis or predicted with a masked language model.
For local arguments, we look for a connecting preposition between the predicate and the argument by inspecting the original sentence or the declarative form of the QA pair.
Otherwise, we use a masked language model \cite{devlin-etal-2019-bert} to rank prepositions given the full passage and the hypothesis.
For more details see \autoref{sec:impl_details}.

\section{Predicate-Argument aware NLI Dataset}
\label{subsec:semantic_aware}
Throughout our experiments, we noticed that the readily available NLI models usually make poor decisions when considering different semantic hypotheses, assigning high probability to propositions with unrelated candidates --- which resonates the findings of \citet{min-etal-2020-syntactic,basmov2023chatgpt}.
We believe that this is caused by the inherent lexical overlap between the hypothesis and the premise texts since our proposition is built entirely from phrases found in the original document.
To circumvent this, we train a semantics-aware entailment model from QA-SRL data.
We use the single-sentence training data and generate entailed and not-entailed propositions.
Each training instance includes a sentence and a proposition centered on a predicate in the sentence.
Positive instances include propositions built using the predicate's argument set.
Each true argument is placed in the hypothesis according to their syntactic position as determined by their QA-SRL question.
The positive propositions are then used to build the negative instances in the following two ways:
The first inserts a noun phrase from the sentence that is not an argument into any position. 
The second switches between syntactic positions of true arguments in the positive proposition, replacing objects as subjects and vice-versa.
This training setup encourages the model to be more sensitive to the semantics of the hypothesis, as encoded in its argument structure.

Our training set contains 465K sentence-hypothesis pairs extracted from the training partitions of QANom \cite{klein-etal-2020-qanom} and QASRLv2~\cite{fitzgerald-etal-2018-large}, with 30\% positive (entailed) instances. 
Negative instances are split between subject-object swaps (14\%), and insertions of non-argument phrases from the sentence (56\%). 
We created multiple positive hypotheses for each predicate by omitting subsets of true arguments, anticipating low coverage conditions of the QA-SRL parser at inference time.
For negative examples, we sampled one positive hypothesis for each predicate and applied our augmentations.

\section{Experiment Setup}
\label{sec:exp_design}
\subsection{Evaluation Datasets}
\label{sec:datasets}
We apply our method to verbal and nominal predicates from several document-level benchmarks. 
\textbf{TNE} \cite{elazar-etal-2022-text}.
We derive our main benchmark from the TNE dataset (\S\ref{sec:background}).
We extract predicate-argument data by focusing on a subset of relations in TNE where the anchor's syntactic head is a deverbal noun, i.e.\ a nominal predicate, and hypothesize that their complements constitute semantic arguments of the predicate word.
To filter the relevant anchors, we apply the nominal predicate classifier of QANom \cite{klein-etal-2020-qanom} with a threshold of 0.75 and identify 10946/1315/1206 predicate instances in the train, development, and test partitions respectively.
The average document in TNE is 8 sentences long, where each deverbal anchor is related to 4.5 complement entities on average, and notably, 2.5 of these have the closest mention to the predicate located in a different sentence.
Examining a sample of 50 deverbal anchors we find that out of 275 cross-sentence complement entities, 93\% exhibit a semantic relation that can be captured by a QA-SRL question, validating our initial hypothesis.

The TNE task pre-specifies a list of noun-phrases as candidates in each document\footnote{Note that in an ordinary SRL setup, a candidate list is not provided.}, and we are tasked to select out of those the complement phrases for each deverbal anchor in the document.

When using generative methods to form arguments, we consider a specific NP candidate as predicted if it matches one of the generated argument phrases. 
Two phrases match if either they share the same syntactic head or have a high token-wise overlap of above 0.5 Intersection-over-Union (IOU).
Otherwise, any non-overlapping generated phrase is discarded.

\textbf{ON5V} \cite{moor-etal-2013-predicate} We also evaluate our method on ON5V, using the unified set of predicates from both train and development partitions as our evaluation data.
The documents in this dataset have gold coreference annotations, which are necessary for our evaluation protocol (see \S\ref{sec:eval}).
We use cross-fold validation over 4 folds and report average and standard deviation on the test fold.
The search for arguments is limited to a context window of 7 sentences, with 5 preceding and 1 subsequent sentence around the predicate, a scope that was found to be sufficient to locate more than 98\% of all originally annotated arguments.
We use noun phrases and named entities as candidates for cross-sentence arguments, they are extracted using Spacy's \cite{honnibal2020spacy} NER and dependency parsers.
These candidate phrases cover 80\% to 90\% of all cross-sentence arguments found in \citet{moor-etal-2013-predicate} and \citet{gerber-chai-2010-beyond} respectively.

To close the coverage gap (see \autoref{sec:background}) for modifier roles we asked an in-house annotator team to go over the existing data and add any argument phrase that can be captured by a QA-SRL question.
The resulting dataset has 3271 arguments with 1800 novel cross-sentence mentions that did not belong to any previously annotated entity, emphasizing the need for exhaustive annotation.
We refer to \autoref{sec:on5v_annot} for more details regarding the annotation protocol.

\subsection{Evaluation}
\label{sec:eval}
We follow the evaluation methodology proposed by \citet{ruppenhofer-etal-2010-semeval} and adapt it to our schema-free task setting.
It states that credit for a semantic argument should be assigned only once, irrespective of the number of multiple mentions it has in the reference or system output.
Since a salient argument can dominate the argument set with multiple mentions, it becomes imperative to disentangle the evaluation of argument detection from co-reference resolution.

Specifically, our evaluation procedure employs externally provided co-reference data and maps predicted and reference arguments to their entity clusters.\footnote{Our evaluation benchmarks provide gold co-reference chains.}
An entity is considered as predicted if it has at least one mention in the system output, and likewise for an entity in the reference set.
We calculate the standard precision and recall metrics over entities, summing entity counts over all instances in the dataset.

A predicted or reference argument mention is mapped to a co-reference cluster if its match score is above 0.5 with one of the mentions in the cluster.
The score between two spans in the document is defined as
\resizebox{\columnwidth}{!}{
$\mathrm{score}(a, m) = \max\{\mathbb{I}{[\mathrm{h}(a) == \mathrm{h}(m)]}, \mathrm{IOU}(a, m)\}$
}
where $\mathrm{h}(a)$ is the index of the syntactic head token\footnote{Head indices are retrieved with the spaCy dependency parser \cite{honnibal2020spacy}} of $a$ and $\mathrm{IOU}$ is the tokenwise intersection over union score.
If an argument is matched to multiple clusters, we select the one with the higher match score.  
Arguments that do not correspond to any pre-existing cluster form a new singleton cluster.\footnote{The evaluation procedure maps matching system and reference arguments to the same singleton cluster where necessary.} 
A pseudo-code of the protocol is provided in \S\ref{sec:eval}

This evaluation procedure measures unlabeled argument detection.
Evaluating labeled accuracy is challenging since our method does not produce explicit labels, but rather provides some signal about the semantic role through the generated hypothesis.
Moreover, our evaluation benchmarks use different label sets that cannot be mapped easily.
Instead, we suggest an analysis in \autoref{sec:analysis} that provides a measure of label accuracy.

\subsection{Baselines}
\label{sec:baselines}
\textbf{NP-SpanBERT} \cite{elazar-etal-2022-text} is a classification model over anchor-complement pairs that was trained directly over TNE, and  based on SpanBERT-Large \cite{joshi-etal-2020-spanbert}.
We apply the classifier on pairs of deverbal anchors and any other NP in the document and consider the phrase as an argument if the predicted label is any valid preposition.

\textbf{QA-SRL Parser} We re-train the generative parser from \citet{klein-etal-2022-qasem} over a joint training set consisting of sentence level QA-SRL annotations for verbal and nominal predicates \cite{fitzgerald-etal-2018-large, klein-etal-2020-qanom} using a T5-Large encoder-decoder \cite{raffel-etal-2020-exploring}.
The parser is trained over examples of a sentence and a marked predicate word as input and produces questions and answers in the QA-SRL format in its output, where each answer is a semantic argument.
Our re-trained parser has significant performance boosts vs.\ previous published models on the QA-SRL data, for details refer to \autoref{sec:qasem_parser}.
Training is performed for 5 epochs until convergence, using the Adam optimizer with a learning rate of $5e-05$ and a batch size of 16.

For the baseline, we simply apply the parser over complete passages during inference.

\textbf{TNE-Parser} Re-using the joint QA-SRL setup \cite{klein-etal-2022-qasem}, we train a parser directly over passage-level TNE data over the deverbal subset of anchors.
The parser takes a passage with the marked anchor (the head word) as the predicate and outputs questions and answers.
Questions are encoded using the "[anchor] [preposition]?" template to signify the semantic relation between the pair, e.g. \textit{"investigation by?"} and the answer is the complement-argument phrase of that relation.

\textbf{Mistral}
We evaluate a prompting approach using the open-source Mistral-7B (v0.1) instruction-tuned model \cite{jiang2023mistral}.
We design two different prompts for the task, each includes an instruction, a few examples (2-5) in the required format, and the passage with the predicate surrounded by special tags.
The first prompt variant (\textbf{Arg}) asks the model to produce a list of semantically related arguments of the marked predicate, while the second variant (\textbf{QA}) asks for a combined representation of an argument and its semantic role represented as a natural language question-answer pair (refer to \autoref{sec:prompt_examples} for concrete prompt examples).
For ON5V we use examples from the QASRL-GS development set \cite{roit-etal-2020-controlled} containing a high ratio of implicit arguments.
For TNE, we use examples from the TNE training set, with questions formatted in the TNE-Parser format.
The examples are randomly selected and kept fixed for the entire evaluation, to reduce the dependence on specific examples we repeat the evaluation four times and report the average and standard deviation.
Decoding is performed with beam-search, beam width is set to 4.

\begin{table*}[ht!]
\centering
\resizebox{\textwidth}{!}{%
\begin{tabular}{@{}llllllllll@{}}
\toprule
                                       &              & \phantom{aaaa} & \multicolumn{3}{c}{Full Document} &&\multicolumn{3}{c}{Cross-Sentence} \\
\cmidrule{4-6} \cmidrule{8-10}
System                                 & Training Data && Precision & Recall  & F1 && Precision & Recall  & F1      \\ \midrule
\emph{Baselines} \\
NP-SpanBERT \footnotesize{(LG)} & {TNE}         && 75.33 & 42.86 & 54.63 && 66.46 & 36.60 & 47.20 \\
TNE-Parser \footnotesize{(T5-LG)} & {TNE}         && 62.60 & 51.73 & 56.65 && 51.57 & 40.02 & 45.07 \\
QA-SRL Parser \footnotesize{(T5-LG)} & {QA-SRL}      && \textbf{84.77} & 25.14 & 38.77 && \textbf{79.85} &  7.64 & 13.95 \\
Mistral \footnotesize{(Arg, 7B)} & {Instructions}         && 35.62\tiny{$\pm$7.3} &  52.93 \tiny{$\pm$14.9} &  40.72\tiny{$\pm$2.1} && 26.76 \tiny{$\pm$ 6.2} &  48.81 \tiny{$\pm$15.2} &  32.70\tiny{$\pm$1.2} \\
Mistral \footnotesize{(QA, 7B)} & {Instructions}         && 46.29\tiny{$\pm$3.5} &  18.03\tiny{$\pm$3.9} &   25.85\tiny{$\pm$4.4} && 34.95 \tiny{$\pm$3.7} & 15.50\tiny{$\pm$3.0} &   21.41\tiny{$\pm$3.3}  \\
\midrule
\emph{Entailment-based models} \\
Instruct-NLI \footnotesize{(Mistral 7B)} & {Instructions}&& 49.22 & 53.55 & 51.29 && 36.01 & 41.49 & 38.56\\
NLI \footnotesize{(DeBERTa LG)} & {NLI mix.}    && 47.42 & 58.76 & 52.49 && 36.09 & 49.37 & 41.70 \\
SRL-NLI \footnotesize{(DeBERTa LG)} & \small{NLI mix. + QA-SRL} && 56.52 & \textbf{60.29} & \textbf{58.34} && 46.41 & \textbf{50.43} & \textbf{48.34} \\

\bottomrule
\end{tabular}%
}

\caption{Results on the TNE test set for argument detection. Metrics are entity-level --- multiple mentions of the same entity are considered as one. ``Full Document'' refers to results evaluated on all of the arguments, while ``Cross-Sentence" considers only those reference and predicted arguments that have their closest mention to the anchor predicate appear in a different sentence. Direct prompting methods (Mistral) results include standard deviation (SD) over 4 runs with different examples.}
\label{tab:main_results}
\end{table*}
\begin{table}[ht!]
\centering
\resizebox{\columnwidth}{!}{%
\begin{tabular}{@{}lllll@{}}
\toprule
System                         && Precision            & Recall    & F1       \\ \midrule
\emph{Baselines} \\
QA-SRL Parser \footnotesize{(T5-LG)}             && \textbf{58.33}      & 1.29      & 2.52      \\ 
Mistral \footnotesize{(Arg, 7B)}              &&  9.93\tiny{$\pm$3}    & 20.24\tiny{$\pm$6.6} & 12.71\tiny{$\pm$2.1} \\
Mistral \footnotesize{(QA, 7B)}               &&  7.04\tiny{$\pm$1.4}  & 11.41\tiny{$\pm$1}	 &  8.60\tiny{$\pm$0.9} \\
\midrule
\emph{Entailment-based models} \\
Instruct-NLI \footnotesize{(Mistral 7B)}               && 16.34\tiny{$\pm$1.1} & 39.47\tiny{$\pm$5.4} & 22.99\tiny{$\pm$1} \\
NLI \footnotesize{(DeBERTa-LG)}               && 16.90\tiny{$\pm$2.4} & \textbf{52.13}\tiny{$\pm$3.3} & 25.47\tiny{$\pm$3} \\
SRL-NLI \footnotesize{(DeBERTa-LG)}                   && 25.41\tiny{$\pm$5.5} & 36.10\tiny{$\pm$5.2} & \textbf{29.28}\tiny{$\pm$3} 
\\
 \bottomrule
\end{tabular}%
}

\caption{Results on the ON5V unified evaluation set on \textit{cross-sentence} arguments (see Appendix \ref{sec:on5v_full} for Full Document results). We evaluated only those reference and predicted arguments that their closest mention to the predicate appears in a different sentence. All NLI methods use cross-fold validation of 4 folds to determine the classification threshold and report mean and SD over the test folds. Direct prompting methods (Mistral) report mean and SD of 4 runs with different sets of examples.} 
\label{tab:on5v_results}
\end{table}

\subsection{Our Method}
Our entailment-based approach is applied using fine-tuned or zero-shot NLI models.
All models are tuned on the development set for TNE, or using cross-fold validation for ON5V, to find the best-performing classification threshold for candidate phrases.
We use the same NLI model for both phases of local argument and non-local candidate verification.
Noteworthy, the premise in the non-local case is significantly longer, but is limited to the scope of the search, which is 7-8 sentences on average. 
This scope is within the reach of NLI models trained on contemporary datasets as evident in several related works \cite{honovich-etal-2022-true-evaluating, schuster-etal-2022-stretching}.

\textbf{NLI} We use an off-the-shelf NLI model\footnote{\url{https://huggingface.co/MoritzLaurer/DeBERTa-v3-large-mnli-fever-anli-ling-wanli}} based on DeBERTA-V3-Large \cite{he2021debertav3,laurer-etal-2024-less} and trained over a mixture of challenging NLI datasets \cite{parrish-etal-2021-putting-linguist,williams-etal-2018-broad,nie-etal-2020-adversarial,liu-etal-2022-wanli}.
Reported performance is on par with current leading models on MNLI and ANLI. 

\label{subsec:srl-nli}
\textbf{SRL-NLI} We fine-tune our NLI model using the predicate-argument aware dataset (\autoref{subsec:semantic_aware}) with the same architecture as the aforementioned NLI model and initialized to the same weights.
Our model is trained for 3 epochs, with batch size 32 and 5e-6 learning rate.

\textbf{Instruct-NLI} 
We also apply our method with the Mistral LLM serving as the underlying entailment engine.
We assume that the entailment task is embedded in different training regimes and datasets for instruction tuning, and apply the model in a "zero-shot" setting without demonstrating examples in the prompt.
The specific prompt for NLI is re-used from FLAN \cite{wei2022finetuned}, assuming a similar prompt was also used to train Mistral LLM as well.  
We ask for a binary Yes/No answer, where Yes refers to entailment, and verify that one of them is the first emitted token in the response. 
To get a normalized probability of entailment given the premise-hypothesis pair, we apply the softmax function over the corresponding logit values of "Yes" and "No" from the first decoded vector of logits and select the probability of "Yes".

\section{Results}
\label{sec:results}

Tables~\ref{tab:main_results} and~\ref{tab:on5v_results} present the results of the argument detection task on nominal predicates from TNE and verbal predicates from ON5V, respectively. For TNE, we report the results in two settings, (1) \textit{Full Document} considering all semantic arguments in the entire document and (2) \textit{Cross-Sentence}, focusing on arguments located in different sentences than the predicate. This separation allows us to analyze the parsers' performance beyond sentence boundaries. For ON5V, we show results for the Cross-Sentence setting in Table~\ref{tab:on5v_results} and defer Full Document results to Appendix~\ref{sec:on5v_full} due to our focus on cross-sentence performance.

Across both datasets, our predicate-argument-aware entailment model (SRL-NLI), trained on a diverse mix of NLI datasets and further fine-tuned on QA-SRL-derived entailment data (§\ref{subsec:semantic_aware}), exhibits superior overall performance (F1) compared to all evaluated approaches.

\paragraph{Our generic approach outperforms supervised models on TNE}
As shown in Table~\ref{tab:main_results}, our distantly supervised SRL-NLI approach achieves superior performance compared to supervised models like NP-SpanBERT and TNE-Parser, even though these models were directly trained on TNE. This indicates the effectiveness of our approach in tackling semantic argument detection without the need for task-specific supervision.

\paragraph{Predicate-Argument-aware entailment model boost performance} SRL-NLI outperforms NLI (using the same DeBERTa underlying model) by 6.6 F1 points on TNE and 3.8 on ON5V, indicating the benefit of an enhanced classifier that is sensitive to predicate-argument semantics.
\vspace{-1pt}
\paragraph{Cross-sentence is more difficult}
When evaluated on TNE, all examined models undergo a performance deterioration for the more challenging setting of cross-sentence argument detection.
The drop in performance is especially detrimental for the QA-SRL Parser (-24.8 F1), which can be attributed to its single-sentence training scope. 
Notably, NLI-based models exhibit an on-par performance decrease with the TNE parser, which was supervised over task-specific document-level data. Hence, it seems that our SRL-NLI approach enjoys the best of both worlds --- it learns document-level semantic understanding from NLI, while specializing in predicate-argument semantics due to the sentence-level QA-SRL supervision.

\paragraph{LLMs: Simple wins, complex stumbles}

Directly asking Mistral in the few-shot setting to identify all semantic arguments of a predicate within a paragraph leads to subpar performance (40.72 vs.\ 58.34 F1 on TNE and 12.71 vs.\ 29.28 F1 on ON5V for the best Mistral configuration).
Interestingly, prompting Mistral just for the arguments (\textbf{ARG}) consistently achieves higher performance on both TNE and ON5V, than asking it to produce arguments in the form of QA pairs (\textbf{QA}), which could have been more fitting for an instruction tuned model.

However, our approach of framing implicit argument identification as a series of entailment decisions, and leveraging Mistral as a zero-shot entailment model (Instruct-NLI) already yields remarkable performance gains.
This method surpasses directly prompting Mistral for arguments, achieving a 5.9 F1-score improvement on TNE and an impressive 10+ F1-score increase on ON5V.

These results highlight the benefit of decomposing complex tasks into simpler binary decisions for LLMs, potentially due to reduced reasoning burden and better alignment with their instruction fine-tuning data.

\section{Analysis}
\label{sec:analysis}

Our evaluation against the TNE datasets measures unlabelled argument detection, which leaves the role assignment accuracy of our system unexplored. 
Since our approach is schema-independent, the argument's semantic role is not provided explicitly but is expressed through its syntactic position in the proposition. 
We thus tap into the \textit{labeling accuracy} of our system through a manual analysis.  
Specifically, we sample 50 deverbal nominal predicates from the TNE test set along with their 260 gold cross-sentence complements and inspect the complements' highest-ranked proposition during inference.
Each proposition contains the complement in its most probable syntactic position as ranked by our SRL-NLI model.
In order to align the setting of our analysis to a typical use case scenario of our method, we further run an OntoNotes parser \cite{shi2019simple} over the selected propositions to attain  PropBank labels of the arguments.
An author of this paper then verified that the predicted semantic role label matches in definition against the semantic relation captured by TNE annotators. 

Omitting 14 TNE complements that don't correspond to verbal arguments, and 20 arguments that are missed by the OntoNotes parser, the extracted role is accurate at 161/226 (71\%) of the cases.
Mistakes include both OntoNotes parsing mistakes, as well as erroneous syntactic positions selected by the NLI-based ranking.     
For further analysis refer to \autoref{sec:ext_analysis}.

\section{Conclusions}
\label{sec:conclusions}
We have demonstrated how to reformulate the problem of argument detection into an entailment task, and successfully used it to detect arguments across sentence boundaries where training data is scarce. 
Moreover, we have explicated the meaning of these distant arguments in the form of simple and easy-to-grasp propositions that keep the correct semantic role information without committing to a specific label schema.
Our proposed method can augment any SRL or event-extraction schema with cross-sentence arguments at test time, without additional annotation or training. Given a sentence-level parser, one can apply it to the extracted proposition to get a label for the captured argument. 
The propositions can potentially serve applications that require information decomposition into smaller units, e.g. SCUs \cite{nenkova-passonneau-2004-evaluating} for the summarization task and many more.

\section*{Limitations}
First, our method relies on a robust entailment engine that is sensitive to the syntactic argument structure of the hypothesis and has a strong comprehension of the passage. 
As we have discovered, this is not a trivial task even for contemporary NLI models.

Secondly, our method might be prone to correct but undesired entailment judgments. 
For example, when a passage describes several different events with lexically similar predicates (e.g. two acquisition events), we might construct a hypothesis that is correctly entailed due to one event, but incorrect with respect to the target event. 
This problem is inherent to the entailment task. 
Entailment is not conditioned on a specific event in the premise but rather verifies the hypothesis against all the information in the premise text.
We tried to address this issue by incorporating the candidate phrase into a hypothesis with other local arguments of the target event, yet this is not a foolproof method.

Lastly, this method may seem computationally intensive, as every candidate phrase in the document is used for entailment multiple times.
However, we have seen in practice that our method is quick to run even on modest accelerators.
Each classification decision applies a single forward pass in an encoder network, and the number of forward steps is bounded by the number of candidates we examine.
In contrast, a generative approach makes a forward pass at inference time for each \emph{token} of a predicted argument.

\section*{Acknowledgements}
We would like to thank our diligent reviewers for their constructive suggestions and comments.
This work was supported by the Israel Science
Foundation (grant no. 2827/21), and a grant from
the Israel Ministry of Science and Technology.

\bibliography{custom}

\appendix

\section{Evaluation Procedure}
\label{sec:evaluation_impl}
The following excerpt is an example of a partial Python implementation of our evaluation procedure described in \S\ref{sec:eval}. It counts the number of true positive, false positive, and false negative entities in a predicted instance.
\begin{minted}[breaklines]{python}
def count_entities(
  refs, preds,
  all_mentions
) -> tuple[int,int,int]:
  for ref in refs:
    all_mentions = map_cluster(
      ref, all_mentions
    )
  for pred in preds:
    all_mentions = map_cluster(
      pred, all_mentions
    )
  pred_ccs = {a['cc_id'] for a in preds}
  ref_ccs = {a['cc_id'] for a in refs}
  n_tp = len(pred_ccs & ref_ccs)
  n_fp = len(pred_ccs - ref_ccs)
  n_fn = len(ref_ccs - pred_ccs)
  return n_tp, n_fp, n_fn

THRESH = 0.5
def map_cluster(arg, all_mentions):
  score, idx = best_match(
    arg, all_mentions
  )
  if score >= THRESH:
    cc_id = all_mentions[idx]['cc_id']
    arg['cc_id'] = cc_id
  else:
    cc_id = new_cc_id(all_mentions)
    arg['cc_id'] = cc_id
    all_mentions.append(arg)
  return all_mentions

def best_match(arg, all_mentions):
  scores = [
    score(ref, m) for m in all_mentions
  ]
  score, idx = argmax(scores)
  return score, idx
\end{minted}
The input is a list of predicted and reference arguments, and a list of known co-reference mentions.
Each argument and mention item have a start and end token index and also a head token index. 
The mentions also have an entity identifier (cc\_id).

\section{ON5V Results}
\label{sec:on5v_full}
\begin{table}[ht!]
\centering
\resizebox{\columnwidth}{!}{%
\begin{tabular}{@{}lllll@{}}
\toprule
System                         && Precision            & Recall    & F1       \\ \midrule
\emph{Baselines} \\
QA-SRL Parser \footnotesize{(T5-LG)}          &&  89.38 & 37.48 & 52.81 \\ 
Mistral \footnotesize{(Arg, 7B)}              &&  17.01\tiny{$\pm$4.9} & 21.45\tiny{$\pm$5.4} &	18.16\tiny{$\pm$1.7} \\
Mistral \footnotesize{(QA, 7B)}               &&  15.37\tiny{$\pm$3.6} & 15.49\tiny{$\pm$1.1} & 15.27\tiny{$\pm$1.9} \\ \midrule
\emph{Entailment-based models} \\
Instruct-NLI \footnotesize{(Mistral 7B)}      &&  33.97\tiny{$\pm$1.5} & 61.41\tiny{$\pm$4.5} & 16.34\tiny{$\pm$1.1} \\
NLI \footnotesize{(DeBERTa-LG)}               &&  31.28\tiny{$\pm$1.6} & 69.01\tiny{$\pm$2.9} & 43.03\tiny{$\pm$1.9} \\
SRL-NLI \footnotesize{(DeBERTa-LG)}           &&  46.59\tiny{$\pm$7.9} & 61.49\tiny{$\pm$2.4} & 52.64\tiny{$\pm$4.1} \\ \bottomrule
\end{tabular}%
}
\caption{Results on the ON5V unified evaluation set on \textit{full-document} evaluation. All NLI methods use cross-fold validation of 4 folds to determine the classification threshold and report mean and std. dev. over the test folds. Direct prompting methods report an average and std. dev. of 4 runs with different sets of examples.}
\label{tab:on5v_full_results}
\end{table}
For completeness, we add the results for the full document evaluation on ON5V.
We achieve comparable results to the QA-SRL parser on the full document.
The parser does not extract almost any cross-sentence arguments, and its overall results stem from its high in-sentence performance.

\section{ON5V Annotation}
\label{sec:on5v_annot}

\begin{figure*}[t]
    \centering
    \includegraphics[width=\textwidth]{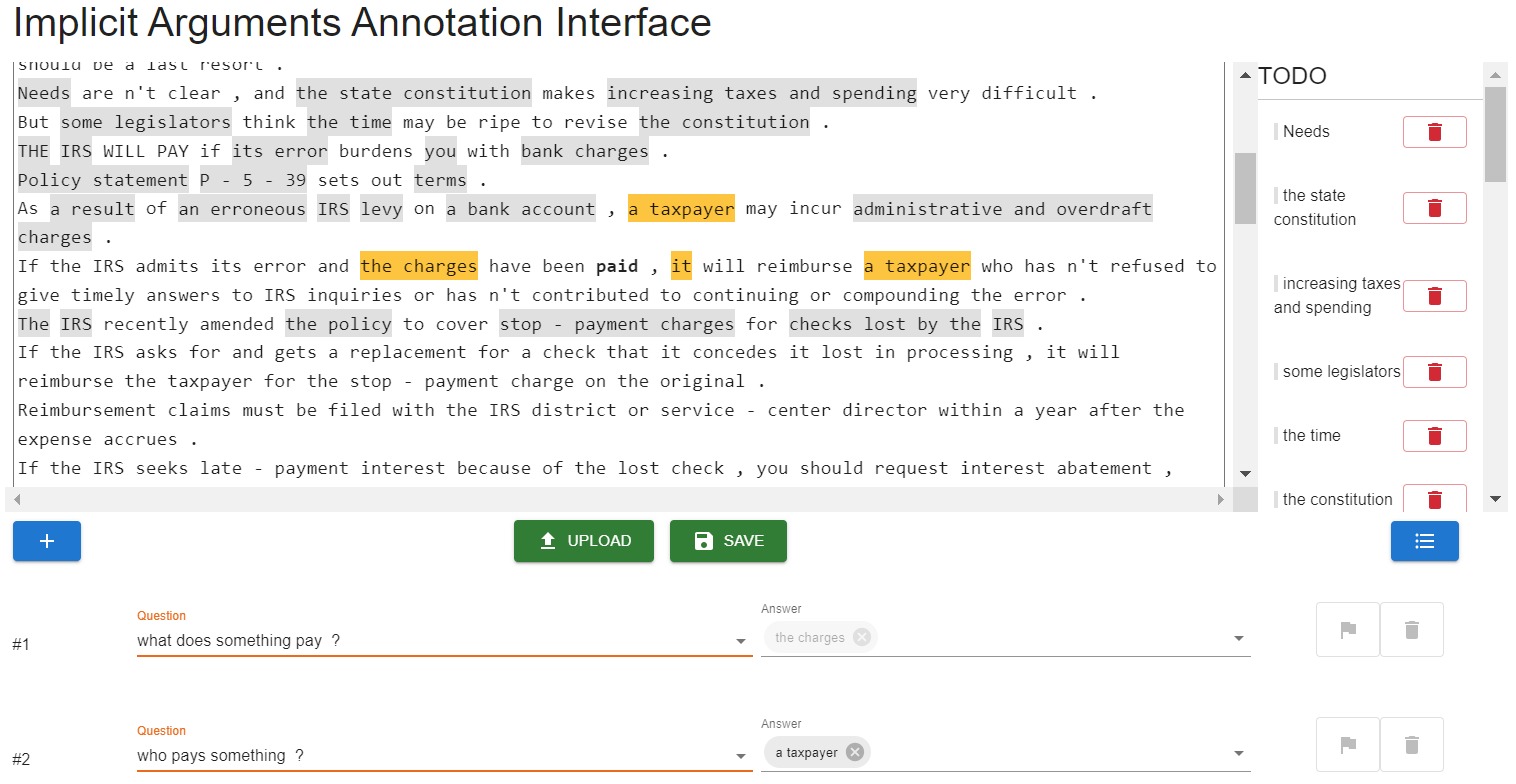}
    \caption{Our implicit arguments annotation interface. The yellow highlighted phrases depicts the current set of arguments, phrases in grey are candidates that need to be either removed from the TODO list or selected as an answer to a QA-SRL question. The interface validates that the question is formatted correctly.}
    \label{fig:annot_interface}
\end{figure*}

We annotated additional arguments for the ON5V dataset for the existing predicates in the dataset.
Annotators were instructed to add new argument phrases and write a question for each one using the QA-SRL question format.
Our interface, depicted in \autoref{fig:annot_interface}, presents the full document with the predicate and all of the already marked arguments from OntoNotes \cite{pradhan-xue-2009-ontonotes} and ON5V, and a selection of candidate phrases.
Annotators were instructed to add new mentions, and not to modify existing arguments.
In our experience selecting arguments from a wide candidate list, as also performed in TNE \cite{elazar-etal-2022-text}, streamlines annotation on a long passage and helps the annotator in covering lengthy contexts.

We scoped the annotation into a context window of sentences of 5 preceding sentences and 1 subsequent after the predicate. 
Past works have shown that more than 90\% of all implicit arguments can be found within this window \cite{gerber-chai-2010-beyond}. 
Our phrase candidates include noun phrases extracted using the same procedure we describe in \autoref{sec:exp_design}, and the annotator is asked to remove them from a "TODO" list if they are not an argument or write them a proper QA-SRL question.

We recruited 5 in-house annotators, four with a strong background in linguistics and one native English speaker who excelled on our qualification assignment.
We presented them the QA-SRL annotation guidelines from \citet{roit-etal-2020-controlled}, and conducted a short training round of 10-15 predicates, after which we provided personal and detailed feedback.
Each predicate took on average ~5 minutes to annotate. 
During the annotation period, one of the authors examined 10-20\% of each annotator's workload to verify correctness and proper coverage.
We paid each annotator an hourly rate of ~14\$, and annotation took about 10 minutes per predicate.

\section{Extended Analysis}
\label{sec:ext_analysis}
\begin{figure}[t!]
    \centering
    \includegraphics[width=\columnwidth]{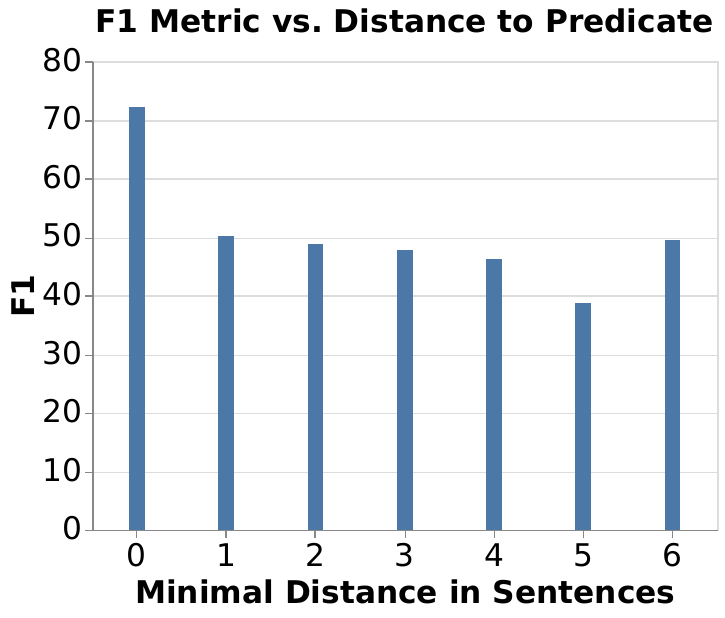}
    \caption{Stratification of our results on the test-set of TNE according to distance counted in sentence between the entity and the predicate. 
    The distance of an entity from the predicate is defined as the absolute difference between the sentence index of the closest mention to the predicate and the sentence index of the predicate.}
    \label{fig:min_dist_no_name_clash}
\end{figure}
To further investigate the performance of our method, we present in \autoref{fig:min_dist_no_name_clash} a stratified view of our TNE test set predictions according to distances from the predicate. 
Since our evaluation protocol counts entities (clusters of argument mentions), to stratify the test set we first group arguments into entities, compute the minimal entity distance to the predicate, and divide accordingly.
Specifically, given a target distance, we select a subset of all predicted and reference arguments whose minimal entity distance to the predicate equals the target value.
We count distances in sentences, where the minimal distance of an entity to the predicate is defined as follows:
\[
\mathrm{MinDistance}(\text{ent}, \mathbf{p}) = \min_{m \in \text{ent}} |\mathrm{sent}(m) - \mathrm{sent}(\mathbf{p})|
\]
where $\mathrm{sent}(a)$ returns the sentence index of the mention. 
All presented strata in the figure have at least 100 gold entities in the evaluated subset.  

The results for in-sentence arguments (first column) can mostly be attributed to the QA-SRL parser.
We observe a slight degradation in performance as the distance of the entity from the predicate increases.


\section{QA-SRL Parser Evaluation}
\label{sec:qasem_parser}
\begin{table}[t!]
\centering
\resizebox{\columnwidth}{!}{%
\begin{tabular}{@{}llllll@{}}
\toprule
System                                & Dataset&& Precision & Recall & F1 \\
T5-Large, retrained                   & Verbal && 91.36 & 64.27 & 75.46 \\
T5-Large, retrained                   & Nominal&& 76.16 & 63.73 & 69.39 \\
T5-Large, retrained                   & ON5V   && 76.48 & 84.35 & 80.22 \\
T5-Small \cite{klein-etal-2022-qasem} & Verbal && 76.20 & 62.40 & 68.60 \\
T5-Small \cite{klein-etal-2022-qasem} & Nominal&& 64.30 & 54.80 & 59.20 \\
 \bottomrule
\end{tabular}%
}

\caption{Results for single sentence evaluation of the re-trained parser on QA-SRL and ON5V evaluation sets.}
\label{tab:qasem_parser}
\end{table}
We re-train the joint QA-SRL parser \cite{klein-etal-2022-qasem} on a T5-Large model and report performance metrics on single sentences. 
Evaluation is conducted with unlabeled mention-level metrics that match spans between reference and predicted arguments.
Results are shown in \autoref{tab:qasem_parser}.
Verbal and Nominal refer to the gold-standard evaluation sets of \citet{roit-etal-2020-controlled} and \citet{klein-etal-2020-qanom} respectively. 
A span match threshold of IOU $>= 0.3$ was used to match previously published metrics.

\section{Prompt Examples}
\label{sec:prompt_examples}
\begin{figure}[t!]
    \centering
    \includegraphics[width=\columnwidth]{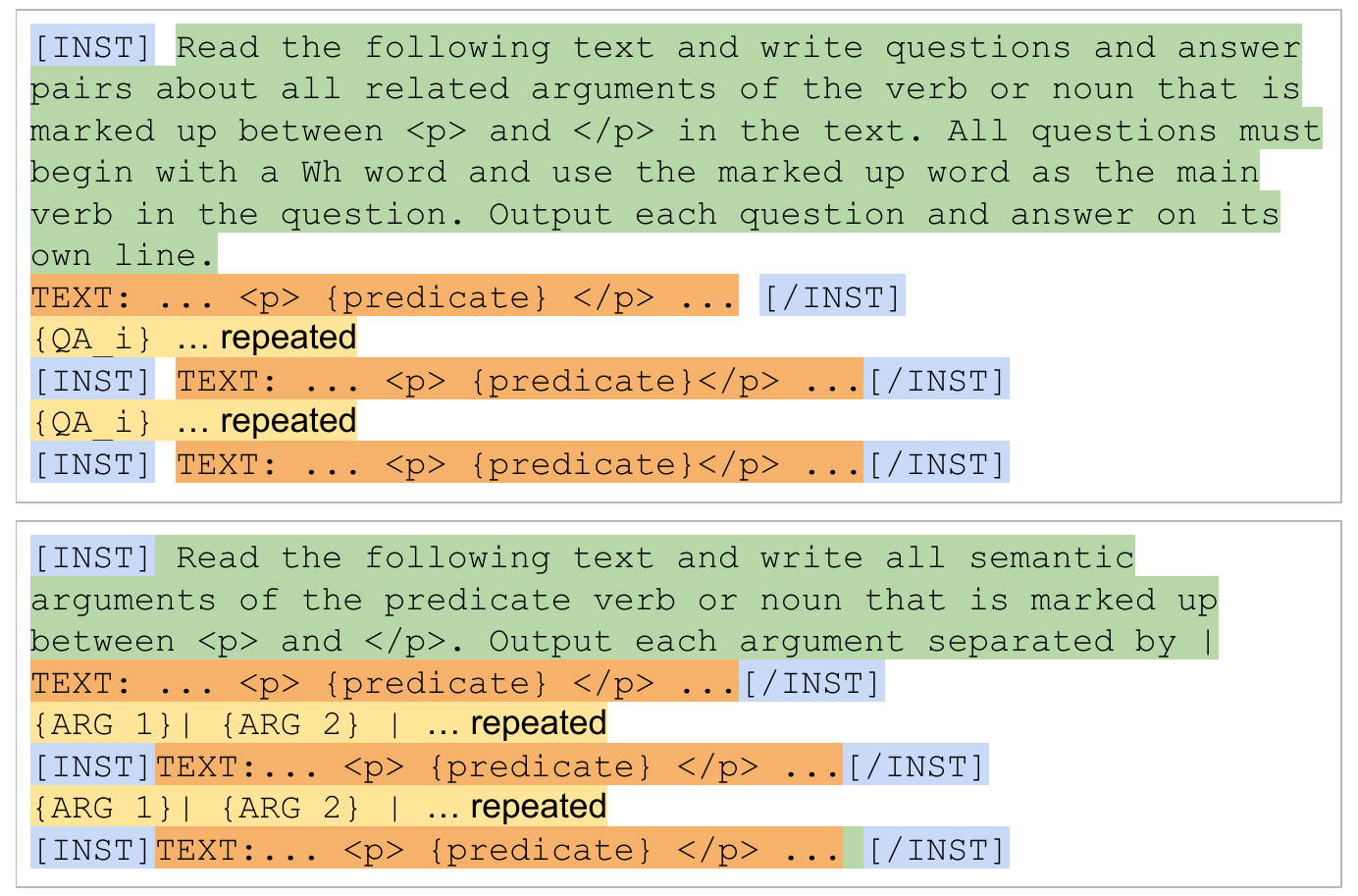}
    \caption{The Mistral-specific prompts are formatted both as QA generation (top) and argument extraction (bottom). Blue highlighting indicates chat instructions, green is our task-specific instruction, orange is for the query, and yellow is our example of a suitable response.}
    \label{fig:chat_qa}
\end{figure}

We provide prompt templates for both the QA prompt and the argument prompt formatted specifically as a chat for the Mistral model in \autoref{fig:chat_qa}.

\section{Implementation Details}
\label{sec:impl_details}
\textbf{Preposition Reranker} 
We use a masked language model \texttt{bert-large-cased} \cite{devlin-etal-2019-bert} to select a preposition for a prepositional phrase (\texttt{ADJ} or \texttt{IOBJ} slots) in the hypothesis.
The input to the language model is a concatenation of the document $\mathbf{D}$ with the hypothesis $\mathcal{H}$.
The hypothesis places \texttt{[MASK]} tokens to indicate the masked preposition token.
We call the language model and extract the output distribution over the vocabulary for the masked token.
Then, we select the highest-ranked preposition by probability out of the following list if it appears in the top 10 ranked words in the vocabulary.
The prepositions include: \textit{on, at, for, to, from, about, as, against, in, with, off, over, into, after, while, before} and \textit{by}. 
Otherwise, we do not prepend a preposition to the phrase.

\textbf{Grammatical Attributes}
The question in QA-SRL can encode different attributes of the event, such as its tense, modality, and negation.
We parse the question of the first local argument in each hypothesis to copy over those attributes from the question to the hypothesis itself.
To determine negation we look for a `\textit{not}' or `\textit{n't}' in the question.
Similarly, to determine modality, we look for the presence of one of the following modal verbs in the auxiliary field of the question: `may', `should', `would', `can', or `might'. 
We implement a heuristic that uses the inflection of the auxiliary and main verb in the question to determine the tense.

\textbf{Local Argument Verificaiton}
All NLI-based methods use the QA-SRL Parser internally to extract local arguments (see \S\ref{sec:structure_from_qasrl}).
Some instances are prone to parsing errors, specifically, question-labels tend to have higher error rates.
This negatively affects our process which is sensitive to the syntactic position of each argument sourced from the question. 
In other cases, more than one local argument may share the same syntactic position.

These problems can be remedied by filtering incorrect arguments and selecting the top-ranked argument based on its entailment score.
We determine the argument's score using the following procedure.
We build \textit{singleton} hypotheses per argument that assign the phrase to a syntactic position according to its question-label.
We also put placeholder arguments (`someone', `something', see \autoref{sec:semantic_hypothesis}) on empty \texttt{SUBJ} and \texttt{DOBJ} fields, creating hypotheses with an intransitive and transitive usage. 
We score the hypotheses using our entailment model with the predicate's sentence as premise, and assign the argument the highest NLI score one of its hypotheses receives. 

It is assumed that if a local argument does not pass a strict NLI threshold then, most likely, it is due to an error associated with its syntactic position (we refer to the high \textit{unlabeled} precision for the parser at \autoref{sec:qasem_parser}).
Hence, the failing argument phrase is appended to the candidate list for further processing and placement in a different syntactic position.
We set a strict threshold for local argument verification of 0.5 for the \texttt{NLI} and \texttt{Instruct-NLI} models and 0.95 for the semantics-aware model \texttt{SRL-NLI}. 
These were determined empirically on a sample of the QA-SRL development set. 

\textbf{Transitive and Intransitive usage}
The valence pattern of a predicate within a sentence can change the semantic interpretation of its arguments.
Consider the differences between: `\textit{Salaries} \textbf{increased} across the sector' and `\textit{The board} \textbf{increased} the salaries across the sector'. 
The subject phrase is interpreted differently given the valence of the predicate.

The valence pattern in our synthetic hypotheses is determined by the presence or absence of the \texttt{DOBJ} argument.
In some cases, the \texttt{DOBJ} slot is left unassigned for various reasons, however, its presence is necessary to interpret the predicate in the correct sense according to the document.
In such cases, we add another hypothesis assigning the abstract placeholder `something' to \texttt{DOBJ} and score it according to the usual flow of our method.

\end{document}